\RequirePackage{xcolor}
\documentclass[journal,twoside,web]{IEEEtran}

\usepackage{cite}
\usepackage{amsmath,amssymb,amsfonts}
\usepackage{algorithm}
\usepackage[noEnd=true, indLines=true, spaceRequire=false, italicComments=false, rightComments=true, commentColor=gray]{algpseudocodex}

\usepackage{graphicx}
\usepackage{textcomp}
\usepackage{booktabs}
\usepackage{multirow,multicol}
\usepackage{bm}
\usepackage{makecell}
\usepackage{url}
\usepackage{siunitx}
\usepackage{subcaption}

\colorlet{shadecolor}{gray!20}

\newcommand{\New}[1]{{\color{black}#1}}

\usepackage{cleveref}
\crefname{figure}{Fig.}{Figs.}
\Crefname{figure}{Fig.}{Figs.}
\crefname{equation}{}{}
\Crefname{equation}{Equation}{Equations}

\providecommand{\Bx}{{\mathbf{x}}}
\providecommand{\By}{{\mathbf{y}}}
\providecommand{\Bxt}{{\mathbf{x}_t}}
\providecommand{\Bxzero}{{\mathbf{x}_0}}

\providecommand{\Normal}{{\mathcal{N}}}
\providecommand{\alphabarT}{{\Bar{\alpha}_t}}
\providecommand{\alphabar}{{\Bar{\alpha}}}
\providecommand{\E}{{\mathbb{E}}}
\providecommand{\Bepsilon}{{\bm{\epsilon}}}
\providecommand{\Tweedie}{{\hat{\mathbf{x}}_0}}
\DeclareMathOperator*{\argmin}{arg\,min}

\makeatletter
\newcommand*\bigcdot{\mathpalette\bigcdot@{.5}}
\newcommand*\bigcdot@[2]{\mathbin{\vcenter{\hbox{\scalebox{#2}{$\m@th#1\bullet$}}}}}
\makeatother

\def\BibTeX{{\rm B\kern-.05em{\sc i\kern-.025em b}\kern-.08em
    T\kern-.1667em\lower.7ex\hbox{E}\kern-.125emX}}
\begin{document}
\title{Steerable Conditional Diffusion for Out-of-Distribution Adaptation \\ in Medical Image Reconstruction}
\author{Riccardo Barbano, Alexander Denker, Hyungjin Chung, Tae Hoon Roh, Simon Arridge, Peter Maass, Bangti Jin, Jong Chul Ye 
\thanks{R.B., A.D. and H.C. have an equal contribution. R.B. acknowledges support from the i4health PhD studentship (UK EPSRC EP/S021930/1).
A.D. acknowledges support from the Deutsche Forschungsgemeinschaft (DFG, German Research Foundation) - Project number 281474342/GRK2224/2 and from EPSRC programme grant EP/V026259/1.
S.A. and B.J. acknowledge support from UK EPSRC grants EP/T000864/1 and EP/V026259/1.
P.M. acknowledges support from DFG-NSFC project M-0187 of the Sino-German Center mobility programme and by the BAB-project PY2DLL (EFRE).
H.C. and J.C.Y. acknowledge support from the National Research Foundation of Korea under the Grant NRF-2020R1A2B5B03001980.}
\thanks{R.B. was with University College London, and is now with Atinary Technologies. 
A.D. was with University of Bremen; he is now with University College London (e-mail: a.denker@ucl.ac.uk). H.C. and J.C.Y. are with Korea Advanced Institute of Science and Technology. T.H.R. is with Ajou University School of Medicine. S.A. is with University College London. P.M. is with University of Bremen. B.J. is with  Department of Mathematics, The Chinese University of Hong Kong. }
\thanks{For the purpose of open access, the author has applied a Creative Commons Attribution (CC BY) licence to any Author Accepted Manuscript version arising.}
}

\maketitle

\begin{abstract}
Denoising diffusion models have emerged as the go-to generative framework for solving inverse problems in imaging.
A critical concern regarding these models is their performance on out-of-distribution tasks, which remains an under-explored challenge.
Using a diffusion model on an out-of-distribution dataset, realistic reconstructions can be generated, but with hallucinating image features that are uniquely present in the training dataset.
To address this discrepancy and improve reconstruction accuracy, we introduce a novel \New{test-time adaptation} sampling framework called Steerable Conditional Diffusion.
Specifically, this framework adapts the diffusion model, concurrently with image reconstruction, based solely on the information provided by  the available  measurement. 
Utilising the proposed method, we achieve substantial enhancements in out-of-distribution performance across diverse imaging modalities, advancing the robust deployment of denoising diffusion models in real-world applications.
\end{abstract}


\section{Introduction}
\IEEEPARstart{D}{eep} learning methods have transformed the field of image reconstruction, delivering state-of-the-art results across various medical imaging tasks \cite{knoll2020advancing,leuschner2021quantitative}. A broad spectrum of approaches have emerged, ranging from supervised deep reconstructors to unsupervised generative priors, see also the reviews \cite{arridge2019solving,ongie2020deep}. In particular, for medical image reconstruction, deep reconstructors, trained on pairs of clean images and measured data, have dominated the scene \cite{Wang2020tomographyreview}.

However, the performance of these models can deteriorate when they encounter data that differs from their training set. This issue is well-documented in magnetic resonance imaging (MRI), where natural distribution shifts, such as changes in scanner type, image contrast or anatomy, can drastically affect the accuracy of deep learning models~\cite{darestani2021measuring,knoll2019assessment,darestani2022test}.

\begin{figure}
    \centering
    \includegraphics[width=0.8\linewidth]{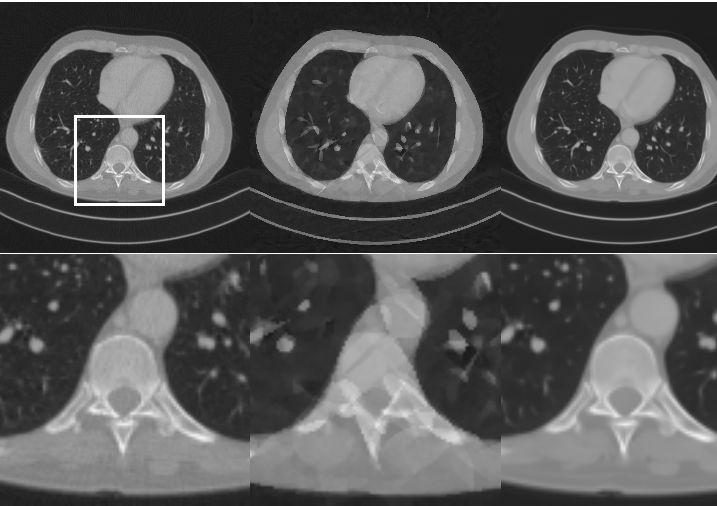}
    
    \caption{Conditional sampling with diffusion models for out-of-distribution data for sparse-view computed tomography with $60$ angles. \textbf{Left}: Ground truth image. \textbf{Middle}: Sample with diffusion model trained on synthetic ellipses. \textbf{Right}: Sample with diffusion model trained on CT images. For the conditional sampling we made use of DDS \cite{chung2024decomposed}, more details in Section \ref{sec:results}. The artefacts in the middle image are due to the mismatch of the ground truth and the training dataset.}
    \label{fig:ood_sampling}
    \vspace{-0.1cm}
\end{figure}

In this work, we focus on medical image reconstruction using denoising diffusion models as generative priors \cite{dhariwal2021diffusion,ho2020denoising,sohl2015deep}. This approach has received increasing attention, with several promising methodologies proposed \cite{song2020score,jalal2021robust,chung2022score,liu2023accelerating,chung2023noise,mardani2024a,wu2023practical,melba2024singh}. Despite their effectiveness in in-distribution tasks, we show that diffusion models face similar challenges as deep reconstructors when applied to out-of-distribution (OOD) data. For instance, as illustrated in \Cref{fig:ood_sampling}, conditional sampling methods can introduce artifacts when faced with distribution shifts. Specifically, the diffusion model was trained on a dataset of synthetic ellipses and then applied to anatomical images. For more details, see Section \ref{sec:results}.
Robustness under distributional shifts and generalising to unseen OOD data is crucial when attempting to reconstruct pathologies that are either underrepresented or entirely absent in the training data.

To this end, we propose a method, named \textbf{S}teerable \textbf{C}onditional \textbf{D}iffusion (SCD), that guides and constrains the generative process to produce images that are consistent with the measured data $\mathbf{y}$.
We achieve this by adapting the diffusion model, concurrently with image reconstruction, based solely on the information provided by a single measurement $\mathbf{y}$. Our contribution can be summarised as follows:
\begin{itemize}
    \item To the best of our knowledge, SCD is the first framework that enables adaptation of diffusion-based inverse solvers for OOD tasks using a single corrupted measured data.
    \item To streamline the adaptation process, we avoid computationally cumbersome fine-tuning of the pretrained network. Instead, we augment the network with a residual pathway using an efficient learnable low-rank decomposition method \cite{hu2021lora}.
\end{itemize}
Finally, our experimental findings show that SCD enhances the image quality across a variety of real-world, OOD imaging reconstruction problems.
This includes sparse-view medical computed tomography (CT), $\mu$CT, volumetric CT super-resolution (SR) and multi-coil MRI. This framework allows for a great flexibility, enabling the utilisation of diffusion models pre-trained on diverse image distributions.

The paper is structured as follows. We start by discussing related work of fine-tuning diffusion models. In Section \ref{sec:background} we cover the necessary background of applying denoising diffusion models to inverse problems in imaging. We present our adaptation in Section \ref{sec:SCD}. After describing the used datasets in Section \ref{sec:datasets}, we present numerical results in Section \ref{sec:results}. Finally, we give a conclusion and outlook for further work.

\subsection{Related Work}
\label{sec:relatedwork}

It has been well documented that deep learning models typically provide worse results if they are evaluated on data that differs from the training set. This has been observed in computer vision tasks such as classification \cite{recht2019imagenet} and in image reconstruction \cite{darestani2021measuring,knoll2019assessment}. However, the scarcity of high-quality paired datasets often necessitates deploying deep learning models in OOD settings. Model adaptation to OOD data is sometimes framed as robustness against distribution shifts~\cite{jalal2021robust}. Closing the performance gap under distribution shifts for supervised deep reconstructors was studied by~\cite{darestani2022test} and~\cite{gilton2021model}. The framework of test-time-training \cite{darestani2022test} studies a similar setting to us and adapts the parameters of a deep reconstructor based on a single available measurement. However, we focus on the adaptation of diffusion models, where test-time training is not directly applicable. 

Recent attempts to enhance the OOD performance of diffusion models, fine-tune a pre-trained diffusion model using a small dataset of ground truth images from the new target domain \cite{abu2022adir,ruiz2023dreambooth,denker2024deft,zhang2023adding}. 
Fine-tuning strategies using low-rank residual pathways while keeping the overall model fixed, were explored in \cite{zhang2023adding,denker2024deft}. However, all of these methods require a paired dataset of images $\Bx$ and measurements $\By$.

Furthermore, methods have been proposed that aim to train a diffusion model from scratch given only corrupted measured data~\mbox{\cite{kawar2023gsure,daras2023ambient}}.
Nevertheless, these methods require two conditions: i) a large collection of data measurements taken under different acquisition protocols; and ii) a full rank condition on the imaging operator \cite{daras2023ambient} or computing its Moore-Penrose pseudo-inverse \cite{kawar2023gsure}. In practice, both conditions are often non-trivial to satisfy.
Concurrently, a similar computational framework, P2L, was proposed to learn the prompt for text-to-image diffusion models~\cite{chung2023prompt}. However, P2L does not adjust the model parameters and hence is incapable of adapting to distinct train-test time distributions. Also, no large-scale open-source text-to-image diffusion models for medical images exist at the time of writing this paper.

\section{Background}\label{sec:background}
\subsection{Medical Image Reconstruction}
Image reconstruction is often posed as an inverse problem with the goal of recovering an image $\Bx\in\mathbb{R}^{d_{x}}$ from (noisy) measured data $\mathbf{y}\in\mathbb{R}^{d_{y}}$, formulated as
\begin{equation}\label{eq:invProbDef} 
   \mathbf{y} =  A\Bx + \boldsymbol{\eta}, 
\end{equation}
where $A\in\mathbb{R}^{d_{y}\times d_{x}}$ is the forward operator, modelling the imaging process. We assume an additive Gaussian noise model, i.e., $\eta\sim\mathcal{N}(\mathbf{0}, \sigma^{2}_{y}I_{d_{y}})$. 
Prior to the advent of deep learning, the recovery of $\Bx$ was often recast using variational regularisation
\begin{equation}\label{eq:invProbOptim}
    \Bx^{*} \in \argmin_{\Bx\in\mathbb{R}^{d_{x}}} \{ \mathcal{L}(\Bx) := \tfrac{1}{2}\|A\Bx - \mathbf{y}\|_{2}^{2} + \lambda \mathcal{R}(\Bx) 
    \},
\end{equation}
where the first term denotes the data-fitting term, and corresponds to the negative log-likelihood of the data $\mathbf{y}$, and the second term, weighted by $\lambda$, is the regularisation functional. Regularisation is needed due to the ill-posedness of the problem. Several imaging reconstructive tasks can be represented via \cref{eq:invProbDef}, where the forward transform $A$ varies with the imaging modality. For instance, Radon transform is used for CT \cite{natterer2001mathematics}, Fourier transform for MRI \cite{natterer2001mathematical}, and down-sampling operator for super-resolution (SR) task. 

In a statistical framework, the regulariser $\mathcal{R}$ can be interpreted as the negative log-likelihood of the prior distribution~\cite{stuart2010inverse}. This opens the door to using deep generative models as data-driven image priors \cite{duff2024regularising}. As denoising diffusion models have shown remarkable abilities in producing high-fidelity images~\cite{song2020denoising,ho2020denoising}, there have been lots of works in using these models as generative priors to solve inverse problems in medical imaging, e.g., \cite{song2020score,song2021solving,chung2024decomposed,jalal2021robust,chung2022score,liu2023accelerating,chung2023noise,mardani2024a,wu2023practical,melba2024singh,feng2023score}.

\begin{figure*}[!t]
    \centering
    \includegraphics[width=0.92\textwidth]{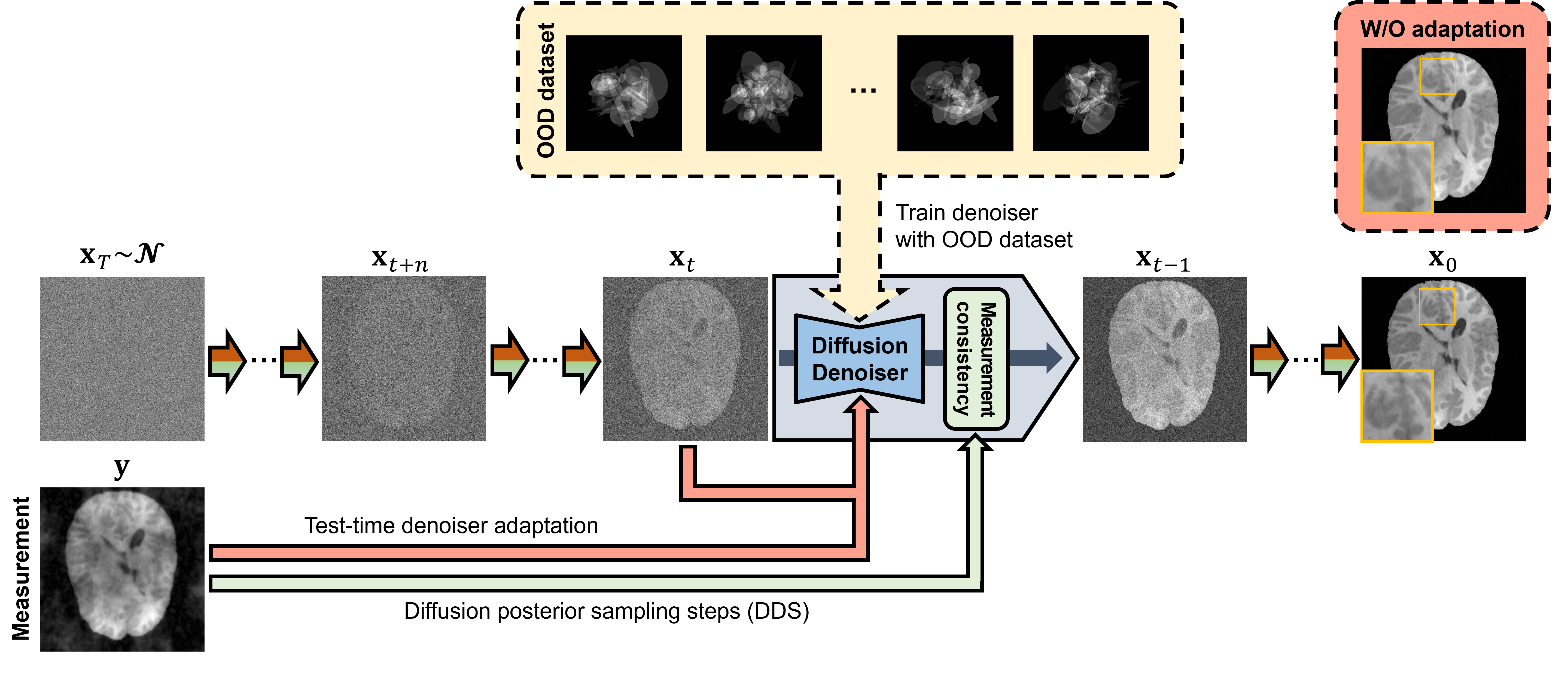}
    \caption{An illustration of the Steerable Conditional Diffusion (SCD) sampling process. In addition to the measurement consistency steps (green), SCD includes an adaptation step (red) to fine-tune the diffusion model on the provided data.}
    \label{fig:flowchart}
    \vspace{-0.2cm}
\end{figure*}
\subsection{Denoising Diffusion Probabilistic Models}
Denoising Diffusion Probabilistic Models (DDPM) \cite{ho2020denoising, sohl2015deep} model the distribution of interest $q(\Bx_{0})$ by constructing a parametric hierarchical model $p(\Bx_{0}; \boldsymbol{\theta}) = \int p(\Bx_{T}; \boldsymbol{\theta})\prod_{t=1}^{T}p(\Bx_{t-1}|\Bx_{t}; \boldsymbol{\theta}) \text{d}\Bx_{1:T}$ with latent variables $\Bx_{\{1, \cdots, T\}}\in \mathbb{R}^{d_{x}}$ and transition densities $p(\Bx_{t-1}|\Bx_{t}; \boldsymbol{\theta})$ with learnable parameters $\boldsymbol{\theta}\in\mathbb{R}^{d_{\theta}}$. This defines a $T$-length parametrised Markov chain, where transitions are learned to reverse a forward conditional diffusion process $q(\Bx_{\{1, \cdots, T\}}|\Bx_{0})$, which gradually adds noise to the data $\Bxzero$ using Gaussian conditional transition kernels defined by
\begin{equation}\label{eq:FwdTransProb}
    \Bxt | \Bx_{t-1} \sim \Normal(\sqrt{1 - \beta_t} \Bx_{t-1}, \beta_t I_{d_{x}}) =: q(\Bxt|\Bx_{t-1}), 
\end{equation}
with forward process variances $\beta_1 < \dots < \beta_T$ for $T$ time steps. Here $\Bx_{0}$ is the noiseless image, and $\Bx_{T}$ is a fully corrupted instance drawn from an easy-to-sample noise distribution, e.g., $\Bx_{T}\sim \mathcal{N}(\mathbf{0}, I_{d_{x}}) =: q(\Bx_{T}) = p(\Bx_{T}; \boldsymbol{\theta})$. The forward process admits closed-form sampling of $\Bx_{t}$ conditioned on $\Bx_{0}$ for all timesteps~\mbox{$t\in\{1, \cdots, T\}$}
\begin{equation}\label{eq:MrgTransProb}
    \Bxt|\Bxzero \sim \Normal(\sqrt{\alphabarT} \Bxzero, (1 - \alphabarT) I_{d_{x}}) =: q(\Bxt|\Bxzero),
\end{equation}
with $\alphabarT = \prod_{i=1}^t (1 - \beta_t)$. 
Training a DDPM amounts to matching the noise 
$\boldsymbol{\epsilon} \sim \mathcal{N}(\mathbf{0}, I_{d_{x}})$ by minimising the so-called $\boldsymbol{\epsilon}$-matching objective
\begin{equation}\label{eq:EpsMatchScore}
    \min_{\boldsymbol{\theta}\in\mathbb{R}^{d_{x}}} \E_{t \in U(\{1,T\}), \Bxzero \sim q(\Bx_{0}), \boldsymbol{\epsilon} \sim \Normal(\mathbf{0}, I_{d_{x}}) }[ \| \Bepsilon(\Bxt, t; \boldsymbol{\theta}) - \boldsymbol{\epsilon} \|_2^2], 
\end{equation}
where $U(\{1,T\})$ denotes the uniform distribution on the set $\{1,\ldots,T\}$ and the noisy sample is represented as $\Bxt = \sqrt{\alphabarT} \Bxzero + \sqrt{1 - \alphabarT} \boldsymbol{\epsilon}$. Moreover, via the $\boldsymbol{\epsilon}$-matching objective, a multi-noise level residual
denoiser $\Bepsilon(\Bxt, t; \boldsymbol{\theta})$ is learned as a proxy for the (Stein) score, i.e., $\nabla_\Bx[\log q](\Bx_{t}) \approx -\boldsymbol{\epsilon}(\Bx_{t}, t; \boldsymbol{\theta}^{*}) / \sqrt{1-\bar{\alpha}}_{t} =: \boldsymbol{s}(\Bx_{t}, t; \boldsymbol{\theta}^{*})$ \cite{song2020score} with $\boldsymbol{\theta}^{*}$ being a minimizer of the $\boldsymbol{\epsilon}$-matching objective. Note that $\nabla_\Bx[\log q](\Bx_{t})$ is the gradient of the log-density of interest with respect to the first argument, evaluated at $\Bx_{t}$. 

The denoiser is trained such that the generative process $p(\Bx_{\{1, \cdots, T\}}; \boldsymbol{\theta}^{*})$ approximates well the intractable reverse process for all $t$.
The reverse diffusion process ends with $\Bx_{0}$ by iteratively denoising a sequence of noisy samples starting from pure noise $\Bx_{T}\sim\mathcal{N}(\mathbf{0}, I_{d_{x}})$. Initially, ancestral sampling was used for solving the reverse process, requiring a large number of time steps \cite{ho2020denoising}. Denoising diffusion implicit models (DDIM)  \cite{song2020denoising} were proposed as an accelerated sampling method. Following DDIM, one can generate a $\Bx_{t-1}$ from $\Bx_{t}$ via,
\begin{align}
	\Bx_{t-1}\!=\!\sqrt{\alphabarT} \Tweedie(\Bxt; \boldsymbol{\theta}^{*})\!+\!\sqrt{1\!-\!\alphabar_{t-1}\!-\!\eta_t^2} \Bepsilon(\Bxt, t; \boldsymbol{\theta}^{*}) + \eta_t \boldsymbol{\epsilon}, 
\end{align}
where we use $\eta_t = \eta \sqrt{(1 - \alphabar_{t-1})/(1 - \alphabar_t)} \sqrt{1 - \alphabar_t / \alphabar_{t-1}}$ and $\Tweedie(\Bxt; \boldsymbol{\theta}^{*})$ is given by Tweedie's formula \cite{efron2011tweedie}, i.e.,
\begin{align}
	\label{eq:Tweedie}
	\Tweedie(\Bxt; \boldsymbol{\theta}^{*}) = (\Bxt - \sqrt{1 - \alphabarT} \Bepsilon(\Bxt, t; \boldsymbol{\theta}^{*})){/\sqrt{\alphabarT}}.
\end{align}
The DDIM update rule consists of three components: the predicted de-noised image, the deterministic noise component, and the stochastic noise component.

\subsection{Diffusion Models in Imaging Problems}
In imaging inverse problems, there is additional conditioning information $\mathbf{y}$, associated to $\Bx_{0}$ via \cref{eq:invProbDef}. The recovery of $\Bx_{0}$ is then conditioned on $\mathbf{y}$, as one aims to embed $\mathbf{y}$ via the posterior distribution of $\Bx_{t}|\mathbf{y}$, or equivalently via its score
\begin{equation}\label{eq:posterior_fact}
	\begin{split}
    \nabla_\Bx[\log p(\bigcdot |\mathbf{y})](\Bx_{t}) &= \nabla_\Bx[\log p](\Bx_{t})\!+\! \nabla_\Bx[\log p(\mathbf{y}|\bigcdot)](\Bx_{t})\\
&\approx \boldsymbol{s}(\Bx_{t}, t; \boldsymbol{\theta})\!+\! \nabla_\Bx[\log p(\mathbf{y}|\bigcdot)](\Bx_{t}),
	\end{split}
\end{equation}
where the prior is approximated using the trained score model.
However, as the score of the likelihood $p(\mathbf{y}|\Bx_{t})$ is only accessible at $t=0$, different sampling strategies have been proposed to approximate this term \cite{song2020score,jalal2021robust,chung2022score,liu2023accelerating,wu2023practical}. A particularly flexible approximation is proposed in Denoising Posterior Sampling (DPS) \cite{chung2023noise}
\begin{equation}
    p(\mathbf{y}|\Bx_{t}) \approx p(\mathbf{y}|\hat{\Bx}_{0}), \text{ with } \hat{\Bx}_{0} := \mathbb{E}_{\Bx_0|\Bxt\sim p(\Bx_0|\Bxt)}[\Bx_{0}|\Bx_{t}], 
\end{equation}
where the posterior mean $\mathbb{E}[\Bx_{0}|\Bx_{t}]$ is computed using Tweedie’s formula~\eqref{eq:Tweedie}. However, as DPS builds on ancestral sampling, a large number of time steps is required, resulting in long sampling time. Alternatively, conditional sampling methods, based on DDIM, were proposed \cite{zhu2023denoising,chung2024decomposed}. 
For instance in DDS \cite{chung2024decomposed}, $\hat{\Bx}_{0}$ is replaced by $\hat{\Bx}'_{0}(\Bxt; \boldsymbol{\theta}^{*}) = \text{CG}^{(p)}(\Tweedie(\Bxt; \boldsymbol{\theta}^{*})),$ where CG takes $p$ conjugate gradient steps with respect to the negative log-density of the measured data defined by \cref{eq:invProbDef} starting with Tweedies estimate as an initialisation. To simplify the notation, we omit the dependence on $\Bx_t$ and $\boldsymbol{\theta}$ where it is not necessary, such as in $\Tweedie$ and $\hat{\Bx}'_{0}$.

\section{Steerable Conditional Diffusion}\label{sec:SCD}
The proposed approach, called \textbf{S}teerable \textbf{C}onditional \textbf{D}iffusion (SCD) sampling, directly adapts the pre-trained diffusion model during the reverse diffusion process based on a single measurement $\mathbf{y}$. As the image $\Bx$ to be recovered is sampled from a distribution of interest $\tilde{q}(\Bx)$ that deviates from the distribution $q(\Bx_{0})$ used in training time, i.e.,~${q}(\Bx) \neq \tilde{q}(\Bx)$, we aim to leverage data consistency to adjust the diffusion model.

However, instead of changing all the weights in the diffusion model, SCD injects additional pathways into the model at an architectural level. These residual pathways are parametrised as low-rank convolutions~\cite{hu2021lora}. Thus, only a small number of parameters are updated and the underlying network is unchanged, reducing the risk of over-fitting. Furthermore, by keeping the original parameters unchanged, we can guarantee that the rich prior learnt from the training data is preserved, and we can always resort back to the original prior by simply turning the residual path off. The SCD pseudo-code is in \Cref{alg:algoSCD} and the flowchart in \Cref{fig:flowchart}.

\begin{algorithm}[t]
\caption{Steerable Conditional Diffusion (SCD)}
\label{alg:algoSCD}
\begin{algorithmic}[1]
    \Require pre-trained diffusion model $\Bepsilon(\Bxt, t; \boldsymbol{\theta}^{*})$
    \Require measured data $\mathbf{y}$, number of sampling steps $T$, number of optim. steps $K$
    \Require adaptation objective $\mathcal{L}$ (cf. \cref{eq:regObjAdapt})
    \Require data-consistency function $\boldsymbol{\Gamma}$ (cf. \cref{eq:conditional_tweedie_proposed})
    \State $\Bx_T \sim \mathcal{N}(\mathbf{0}, I_{d_{x}})$
    \For{$t=T, T-1, \dots, 1$}
         \State ${\boldsymbol{\epsilon}}^{\text{\tiny DDIM}}_{t} \gets\boldsymbol{\epsilon}(\Bx_{t}, t; \boldsymbol{\theta}^{*})$
         \BeginBox[fill=shadecolor]
         \For{$k=1,\dots, K$} \Comment{Adaptation steps}
            \State $\Tweedie \gets (\Bxt - \sqrt{1 - \alphabarT} \Bepsilon(\Bxt, t; \boldsymbol{\theta}^{*}+{\Delta\boldsymbol{\theta}})){/\sqrt{\alphabarT}}$. 
            \State $\hat{\Bx}'_{0} \gets \boldsymbol{\Gamma}(\Tweedie, \By)$
            \State Take gradient descent step on $\nabla_{{\Delta\boldsymbol{\theta}}} \mathcal{L}({\Delta\boldsymbol{\theta}})$ 
         \EndFor 
         \EndBox
         \State ${\boldsymbol{\epsilon}}_{t} \gets \boldsymbol{\epsilon}(\Bx_{t}, t; \boldsymbol{\theta}^{*} + \Delta\boldsymbol{\theta}^{*}_{t})$
         \State $\Tweedie \gets (\Bxt - \sqrt{1 - \alphabarT} \Bepsilon_{t}){/\sqrt{\alphabarT}}$
         \State $\hat{\Bx}'_{0} \gets \boldsymbol{\Gamma}(\Tweedie, \By)$
         \State ${\boldsymbol{\epsilon}} \sim \mathcal{N}(\mathbf{0}, I_{d_{x}})$  
         \State $\Bx_{t-1} \gets \sqrt{\alphabarT}\hat{\Bx}'_{0} + \sqrt{1 - \alphabar_{t-1} - \eta_t^2}{\boldsymbol{\epsilon}}^{\text{\tiny DDIM}}_{t}  + {\eta}_t {\boldsymbol{\epsilon}}$
    \EndFor
    \State \Return $\Bx_0$
\end{algorithmic}
\end{algorithm}

\subsection{Generation-Time Adjustable Parameters Injection}
At generation-time, SCD samples from the reverse diffusion process, augmenting each convolutional layer in $\boldsymbol{\epsilon}(\bigcdot; \boldsymbol{\theta}^{*})$ with learnable low-rank decomposition matrices via Low Rank (LoRA) injection \cite{hu2021lora}. Given a learned matrix $W \in \mathbb{R}^{m \times n}$ representing a convolutional operation in $\boldsymbol{\epsilon}(\bigcdot; \theta^{*})$, the LoRA injection re-writes $W$ as
\begin{equation}
    \tilde{W} = W + \alpha \Delta W, \quad\text{with } \Delta W = B C^\top, 
\end{equation}
where $B \in \mathbb{R}^{m \times r}, C \in \mathbb{R}^{n \times r}$ form a low-rank approximation to the residual update $\Delta W$, i.e., $r\ll \min(m,n)$. \New{The parameter $\alpha \in [0,1]$ controls the strength of the LoRA parametrisation.}
In practice, $B$ is randomly initialised such that all the entries are drawn from a standard Gaussian, while $C$ is set to a zero matrix; thus at initialisation we have $BC^{\top} = \mathbf{0}$. As most diffusion models are implemented as U-Nets, both $B$ and $C$ are implemented as convolutional layers with $r$ output and input channels, respectively. In principle other fine-tuning parametrisation could be considered, e.g., ControlNet \cite{zhang2023adding}. We opt for the LoRA reparametrisation for two primary reasons: i) it introduces an under-parametrised model with concise representations that are robust against overfitting noise in the data \cite{heckel_deep_2018}; and ii) it reduces the memory footprint, meaning it requires less disk space for storage at each adaptation.
We then refer to $\Delta\theta$ as a vector obtained by stacking vectorised elements of $(\Delta W_d)_{d=1}^{D}$, with $\Delta W_{d} = B_{d}C_{d}^{\top}$.

The parameters $\Delta\theta$ of the low-rank residual pathway are trained by minimising the negative log-likelihood at each sampling step, i.e., solving the optimisation problem 
\begin{align}\label{eq:regObjAdapt}
    \Delta \boldsymbol{\theta}^{*} \in \argmin_{\Delta \theta} \tfrac{1}{2} \{ \mathcal{L}(\Delta\boldsymbol{\theta}) := \| A \Tweedie'(\Delta\boldsymbol{\theta}) - \By \|_2^2 
    \}, 
\end{align}
with the conditional Tweedie estimate $\Tweedie'(\Delta\boldsymbol{\theta}) = \E[\Bx_0|\Bx_t, \By]$ depending on $\Delta\boldsymbol{\theta}$. The objective in \eqref{eq:regObjAdapt} is useful since the posterior sampling process with diffusion models is governed by the conditional Tweedie estimate~\cite{peng2024improving}.
In our implementation, the parameters are updated using a small number of update steps with the Adam optimizer~\cite{kingma2014adam}. For the conditional Tweedie estimate, SCD uses the identity~\cite{ravula2023optimizing},
\begin{align}
    \E[\Bxzero|\Bxt, \By] = \frac{\left(\Bxt + (1 - \alphabarT)\nabla_\Bx[\log p(\bigcdot|\mathbf{y})](\Bx_{t})\right)}{\sqrt{\alphabarT}}, 
\end{align}
where the posterior can be factorised according to \cref{eq:posterior_fact}. For the gradient of the prior, we use the approximation with the score model, i.e., $\nabla_\Bx[\log p](\Bx_{t}) \approx \boldsymbol{s}(\Bxt, t; \boldsymbol{\theta}^{*}, \Delta\boldsymbol{\theta}) := -\boldsymbol{\epsilon}(\Bxt, t; \boldsymbol{\theta}^{*}, \Delta\boldsymbol{\theta})/\sqrt{1 - \bar\alpha_t}$~\cite{song2020score}. We propose to approximate the gradient of the likelihood as 
\begin{align}
    -\nabla_{\Bx} [\log p(\By|\bigcdot)](\Bxt) &\approx \frac{\nabla_{\Bx} [\|A\Tweedie - \By\|_2^2](\Bxt)}{2 \sigma_y^2 (1 - \alphabarT) }    
      \\
    &\approx \frac{\sqrt{\alphabarT} \nabla_{\Bx} [\|A\Tweedie - \By\|_2^2](\Tweedie)}{2 \sigma_y^2 (1 - \alphabarT) }    ,
\end{align}
where the first approximation is given by~\cite{chung2023noise}, while the second approximation is similar to what is used in~\cite{chung2024decomposed,chung2023direct}. Notably, the second approximation effectively approximates the Jacobian of the network, i.e., $\partial_{\Bxt} \boldsymbol{\epsilon}(\Bxt, t; \boldsymbol{\theta}^{\ast})$, to be the identity, reducing the computational cost. In a different application, this last approximation is widely used \cite{poole2023dreamfusion}, as the incorporation of the Jacobian is known to be unstable~\cite{du2023reduce,salimans2021should}. Thus, Tweedie estimate conditioned on $\mathbf{y}$ is given by 
\begin{align}
\label{eq:conditional_tweedie_proposed}
    \E[\Bxzero|\Bxt,\By] \approx \Tweedie\!-\!\gamma_t  A^\top\!(\!A\Tweedie - \By) =: \boldsymbol{\Gamma}(\Tweedie, \By) =: \hat{\Bx}'_{0}, 
\end{align}
where $\gamma_t$ is the step-size, incorporating all constants. This approximation is denoted with $\boldsymbol{\Gamma}$, which can alternatively be implemented as any function moving towards the minimizer of the negative log-density of interest, such as one or more steps of gradient descent or CG. In our implementation, we use one step of CG.

In a similar spirit to \cref{eq:invProbOptim}, a regularisation functional can be included in the adaptation process \cref{eq:regObjAdapt}. The addition of additional regularisation will be studied in the sparse-view CT experiments, see Section~\ref{sec:sparseviewct}. Overall, SCD aims to minimise the negative log-likelihood by learning the residual parameters $\Delta\boldsymbol{\theta}$, where the adaptation is repeated at each sampling step~$t$.

\subsection{Sampling from Adjustable Generative Process}

The reverse sampling intertwines the DDIM update rule with the above adaptation step.
SCD's update rule is reported in \Cref{alg:algoSCD}. Note that the augmented network, i.e., $\boldsymbol{\epsilon}(\bigcdot; \boldsymbol{\theta}^{*} +\Delta\boldsymbol{\theta})$ is only used for the predicted de-noised estimate $\Bx_0$, while the deterministic noise update uses $\boldsymbol{\epsilon}(\bigcdot;\boldsymbol{\theta}^{*})$. This is to avoid the ``de-naturalisation'' of $\boldsymbol{\epsilon}(\bigcdot;\boldsymbol{\theta}^{*})$ to act as a multi-noise level denoiser due to the adjustable parameters injection being learned to solve a reconstructive task. This design choice arises as the outcome of the proposed methods, where the log prior score is augmented with a residual architectural pathway. This pathway adjusts the score according to $\mathbf{y}$, ensuring that $\Tweedie'$ is consistent with the measured data $\mathbf{y}$.
\vspace{-0.15cm}

\section{Datasets}\label{sec:datasets}
\label{sec:exp_details}
In the following, we describe the datasets used in the experiments. These datasets allow us to study a wide variety of different distribution shifts, including anatomy shifts (train on brain images and test on knee images) and strong domain shifts (training on synthetic ellipses, testing on abdominal CT scans) as two examples.

\subsubsection{$\mu$CT Walnut \cite{der2019cone}}
The $\mu$CT \textsc{Walnut} dataset includes cone-beam projection data and high-quality reconstructions with a resolution of $501 \times 501$ px. The complete measurements span $1200$ angles and $768$ detector pixels. We subsample these measurements to create sparse-view CT data. The subsampled forward operator is a sparse-matrix operator used to reconstruct the middle slice of the 3D volume \cite{barbano2022educated}. Since the diffusion models under consideration are trained on an image resolution of $256 \times 256$ px, an up-sampling operator is added to the inverse problem. The reconstruction task then becomes recovering $\Bx$, from the measured data $\By$, i.e.,$\By = A(S(\Bx))$, where $S$ describes nearest-neighbour up-sampling to $501 \times 501$ px and $A$ is the sparse-view CT operator.

\subsubsection{AAPM \cite{moen2021low}}
We use a dataset of abdominal CT scans released by the American Association of Physicists in Medicine (\textsc{AAPM}) for the $2016$ grand challenge. 
We transform the data according to \cite{kang2017deep}. 
The \textsc{AAPM} dataset consists of $3839$ training images resized to $256 \times 256$ px. 
We construct a held-out validation set comprising $10$ slices, used for the hyperparameter search, and a held-out test set comprising $56$ images (equidistant with respect to the $z$-axis) to compute performance image metrics.

\subsubsection{LoDoPab Dataset \cite{leuschner2021lodopab}}
The \textsc{LoDoPab} dataset contains human chest CT scans. 
We resize the \textsc{LoDoPab} test set to $256 \times 256$ px and rotate each slice to match the anatomical orientation in \textsc{AAPM}. Analogously, we construct a held-out validation set (out of the test set) comprising $10$ slices, used for the hyperparameter search. Additionally, $100$ slices are taken from the test set to compute performance image metrics.

\subsubsection{Ellipses \cite{adler2018learned}} 
The \textsc{Ellipses} dataset, is a synthetic dataset containing images of a varying number of ellipses with random orientations and sizes \cite{barbano2022educated}. In particular, we generate $32000$ images on-the-fly. The use of synthetic data for training is particularly useful in imaging domain where obtaining large-scale datasets is infeasible or expensive.

\subsubsection{BrainWeb \cite{aubert2006twenty}}
\New{The \textsc{BrainWeb} dataset consists of $20$ realistic simulated 3D volumes. We train the diffusion model on 2D slices. We use the pre-processed data from \cite{schramm_2021_4897350} which includes a simulated $^{18}$F-Fluorodeoxyglucose (FDG) tracer in order to obtain a representative dataset for positron emission tomography (PET). The final training dataset has 4569 slices and is the same as in \cite{melba2024singh}. This dataset is used to evaluate whether SCD can generalise from PET images to CT images.}

\subsubsection{FastMRI \cite{zbontar2018fastmri}}
The FastMRI datasets include data from both clinical \textsc{KNEE} and \textsc{BRAIN} MRI scans. 
The model trained on \textsc{KNEE} used $29877 $ slices from $973$ volumes for training. The model trained on \textsc{BRAIN} used $83216$ slices from $3842$ volumes for training. 
We drop the first/last $5$ slices from each volume as they mostly contain noise. 
The evaluation was done using $236$ slices from $18$ test volumes for \textsc{BRAIN} and $294$ slices from $10$ test volumes for \textsc{KNEE}. We use the masks from \cite{zbontar2018fastmri}, i.e., fully-sample the central region to include $8\%$ of all vertical k-space lines and sample the rest of k-space uniformly at random to reach $4$ fold acceleration.

\subsubsection{BRATS \cite{menze2014multimodal}}
The Multimodal brain tumor image segmentation benchmark (\textsc{Brats}) 2018~\cite{menze2014multimodal} contains 3D brain MRIs that consist of images in different contrast, including T1W, T2W, and FLAIR. We test on 5 T1 contrast and 5 T2 contrast volumes of size $192\times192\times155$. The resulting test set consists of 1550 2D slices. Note that since we use the diffusion model trained on \textsc{Ellipses} phantom dataset, there is no overlap in the training and test sets. For testing, we consider variable density (VD) Poisson disc sampling pattern~\cite{dwork2021fast} ($\times 8$ acceleration) in a single-coil setting.

\subsubsection{$\mu$CT Pig}
A domestic pig head was prepared for $\mu$CT scanning. 
Imaging was conducted using the Skyscan 1273 $\mu$CT scanner (Bruker, Kontich, Belgium). 
A resolution of 100 $\mu$m was chosen based on the specimen's dimensions and imaging needs. The tube voltage was set at 125 kV, with a tube current of 300 $\mu$A. 
A rotation step of 0.3$^\circ$ was selected, with a scanning position of 112 mm, capturing 689 projections. Image reconstruction was performed using the Skyscan NRecon software, and the images were adjusted to enhance visualisation of the specimen's internal structures. 
The resulting images had a resolution of $768\times 768$ px, we used the $512 \times 512$ px center crop from $3768$ slices. In addition, a 5mm coarse-vertical-resolution multi-detector computed tomography (MDCT)~\cite{kalra2004multidetector} scan of the head of a pig is also collected. The pig was laid supine on the CT table, with their head stabilised in a neutral position using a head holder.
Imaging was performed using the Siemens Somatom Force (Siemens Healthineers, Erlangen, Germany).
The tube voltage ranged from 100 to 120 kV, and the tube current was adjusted between 125 and 330 mA. A medium field of view (FOV) was selected to cover the entire brain, reducing peripheral artefacts. 

\begin{figure*}[!ht]
\centering
\includegraphics[width=\textwidth]{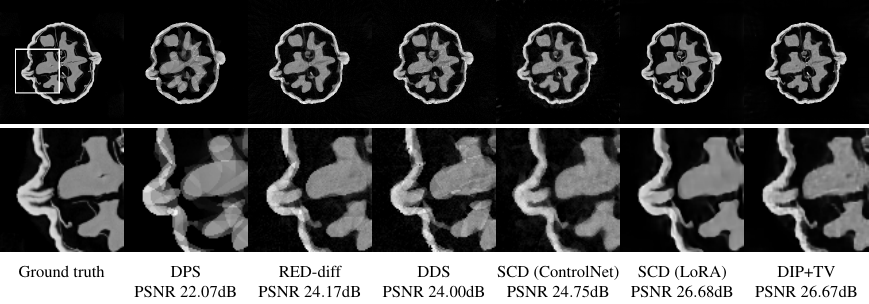}
\caption{DPS, Red-diff, DDS, DIP+TV and SCD (ours) are compared to reconstruct real-measured $\mu$CT data of a walnut from $60$ angles and $128$ detector pixels. The diffusion model was trained on \textsc{Ellipses}. The non-adaptation methods clearly show ellipsoid artefacts in this OOD scenario. \textit{Top:} full image. \textit{Bottom:} zoomed-in part.} \label{fig:results_walnut}
\end{figure*}

\section{Results}\label{sec:results}
We test SCD on sparse-view CT, accelerated MRI and volumetric super-resolution on the datasets described in Section \ref{sec:datasets}. 
All diffusion models are based on the Attention U-Net \cite{dhariwal2021diffusion}. 
We use the LoRA reparametrisation for the attention and convolutional layers. 
Additionally, we retrain all biases in the U-Net following \cite{hu2021lora}, since they only account for a negligible number of parameters. In total, LoRA adds about $0.5-2.5\%$ additional parameters to the diffusion models, depending on the rank $r$. The time-step encoding is not adapted during the fine-tuning process.
For all our experiments, we tune the hyperparameters to maximise the peak signal-to-noise ratio (PSNR) on a validation set. We compute the PSNR and the structural similarity index measure (SSIM) \cite{wang2004image} on a held-out test set. \New{We provide an implementation of our algorithm at \url{https://github.com/alexdenker/SteerableConditionalDiffusion}.}
\vspace{-0.25cm}

\begin{figure}[]
    \centering
    \includegraphics[trim={0 0.25cm 0 0.3cm},clip,width=0.5\textwidth]{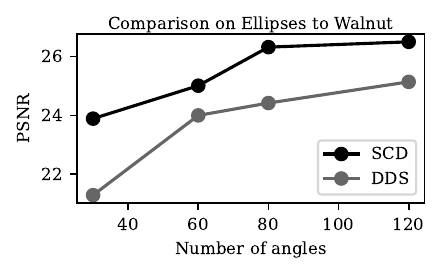}
    \caption{Varying the number of angles for $\mu$CT reconstruction for SCD and DDS from $30$ to $120$ angles. For all experiments, SCD is able to outperform DDS on this OOD task. Both SCD and DDS were tuned for $60$ angles and then applied to the different sparse-view settings.} \label{fig:number_of_anlges}
\end{figure}

\subsection{Initial evaluation on $\mu$CT Walnut}
To highlight the generalisation issues of widely used conditional sampling techniques, we study the reconstruction of  $\mu$CT walnut with a diffusion model trained on the \textsc{Ellipses} dataset. We compare SCD against DPS \cite{chung2023noise}, RED-diff \cite{mardani2024a} and DDS \cite{chung2024decomposed}. For each method, we choose the hyperparameters to maximise the PSNR of the reconstruction. The results are shown in \cref{fig:results_walnut}, where we choose a subsampling of $60$ equidistant angles and $128$ equidistant detector pixels. When using non-adaptation methods, diffusion models trained on the \textsc{Ellipses} dataset exhibit strong artefacts and hallucinations: Small ellipses are clearly visible in the reconstructed walnut. In sharp contrast, SCD results in a robust recovery of the walnut. However, we see that for SCD some minor details of the ground truth are smoothed out. \New{To evaluate the efficiency of the LoRA parametrisation, we employ a version of SCD using ControlNet \cite{zhang2023adding}. In \cref{fig:results_walnut} we see that also with the ControlNet backbone SCD is able to adapt to the new walnut. However, this parametrisation leads to an increase in sampling time, as $38\%$ additional parameters are added in contrast to the 0.5-2.5\% for LoRA.} We also include a deep image prior (DIP+TV) approach \cite{ulyanov2018deep,baguer2020computed}. DIP is well-known to be prone to overfitting, thus we use early stopping based on the highest PSNR obtained during the optimisation process, i.e., reporting an optimistic oracle PSNR.

SCD adapts to a new distribution with a data consistency loss. We evaluate SCD in the scenario, that the likelihood becomes less informative.
For this, we evaluate the performance of SCD against DDS by gradually decreasing the number of angles.
Here, we test subsampling of the measurements to $30, 60, 80$ and $120$ equidistant angles. This corresponds to an undersampling factor of $1.4-5$. The results are presented in~\Cref{fig:number_of_anlges}. We used the same hyperparameters for all settings. Interestingly, even under the extreme case of 30 angles, we observe no significant decrease in the performance of SCD, showing similar advantages of the adaptation process as opposed to using DDS. This suggests that SCD is also useful in the regime where the measurement is highly sparse. %

\begin{figure*}[ht!]
   \centering
   \includegraphics[width=0.95\textwidth]{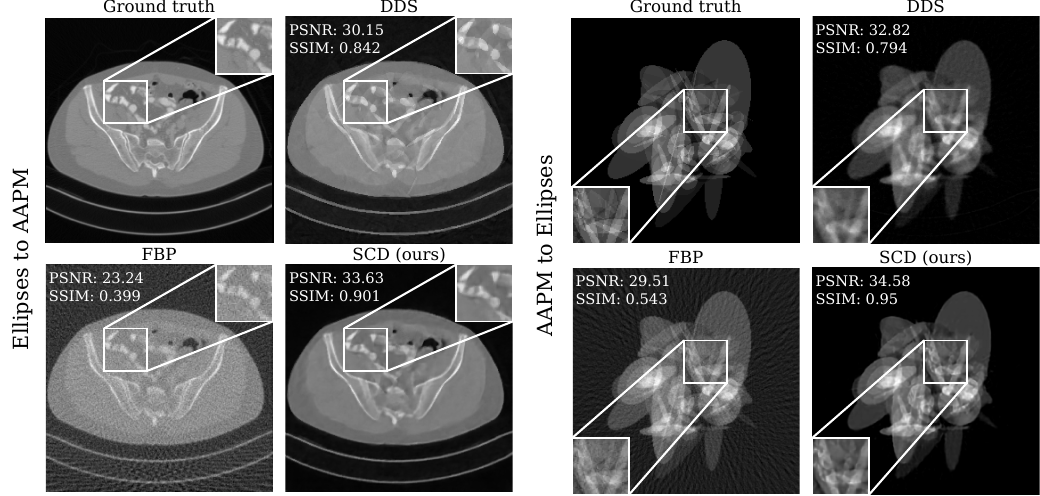}
   \caption{\textbf{Left:} Results for sparse view CT for training on \textsc{Ellipses} and testing on \textsc{AAPM}. \textbf{Right:} Results for sparse view CT for training on \textsc{AAPM} to testing on \textsc{Ellipses}. We compare SCD against DDS and the filtered-back projection.}
   \label{fig:sparseViewCT}
\end{figure*}

\tabcolsep=0.11cm
\begin{table*}[ht!]
\centering
\caption{PSNR and SSIM (mean $\pm$ SE) for SCD vs. baselines on sparse-view CT. \textbf{Bold}: best }\label{tab:resultsSparseViewCT}
\begin{tabular}{llcccccc}
\toprule \multirow{ 2}{*}{$q(\Bx_{0})$} & \multirow{ 2}{*}{$\tilde{q}(\Bx_{0})$}  & \multicolumn{2}{c}{\textsc{RED-diff}}  & \multicolumn{2}{c}{\textsc{DDS}} & \multicolumn{2}{c}{\textsc{SCD (\textbf{ours})}} \\ 
 &  & PSNR  & SSIM & PSNR  & SSIM & PSNR  & SSIM       \\ \midrule
\multirow{ 3}{*}{\textsc{AAPM} } & \textsc{AAPM}  & $38.37$ {\color{gray} \scriptsize $\pm 0.09$} & $0.941$ {\color{gray} \scriptsize $\pm 0.001$}    & $39.55$ {\color{gray} \scriptsize $\pm 0.08$}  & $0.951$ {\color{gray} \scriptsize $\pm 0.001$} & $\mathbf{39.73}$ {\color{gray} \scriptsize $\pm 0.07$} & $\mathbf{0.952}$ {\color{gray} \scriptsize $\pm 0.001$} \\
 & \textsc{LoDoPab}  & $31.09$ {\color{gray} \scriptsize $\pm 0.18$} & $0.799$ {\color{gray} \scriptsize $\pm 0.006$} & $31.20$ {\color{gray} \scriptsize $\pm 0.21$}  & $0.742$ {\color{gray} \scriptsize $\pm 0.007$} & $\mathbf{34.21}$ {\color{gray} \scriptsize $\pm 0.25$} & $\mathbf{0.850}$ {\color{gray} \scriptsize $\pm 0.008$} \\
 &\textsc{Ellipses} & $32.67$ {\color{gray} \scriptsize $\pm 0.10$} & $0.772$ {\color{gray} \scriptsize $\pm 0.006$} & $33.11$ {\color{gray} \scriptsize $\pm 0.11$} & $0.787$ {\color{gray} \scriptsize $\pm 0.005$} & $\mathbf{35.41}$ {\color{gray} \scriptsize $\pm 0.10$} & $\mathbf{0.954}$ {\color{gray} \scriptsize $\pm 0.001$} \\ \cmidrule{3-8} 
\multirow{ 3}{*}{\textsc{Ellipses} } & \textsc{Ellipses}  & $35.51$ {\color{gray} \scriptsize$\pm 0.09$} & $0.878$ {\color{gray} \scriptsize$\pm 0.004$} & $\mathbf{36.91}$ {\color{gray} \scriptsize$\pm 0.09$} & $\mathbf{0.972}$ {\color{gray} \scriptsize$\pm 0.001$} & $36.02$ {\color{gray} \scriptsize$\pm 0.08$} & $0.968$ {\color{gray} \scriptsize$\pm 0.000$} \\
& \textsc{AAPM} &$29.75$ {\color{gray} \scriptsize $\pm 0.06$} & $0.801$ {\color{gray} \scriptsize $\pm 0.002$} & $30.82$ {\color{gray} \scriptsize$\pm 0.06$}  & $0.847$ {\color{gray} \scriptsize$\pm 0.002$} & $\mathbf{33.98}$ {\color{gray} \scriptsize$\pm 0.12$} & $\mathbf{0.883}$ {\color{gray} \scriptsize$\pm 0.002$} \\ 
& \textsc{LoDoPab} & $31.25$ {\color{gray} \scriptsize$\pm 0.22$} & $0.742$ {\color{gray} \scriptsize$\pm 0.008$}&$31.80$ {\color{gray} \scriptsize$\pm 0.22$}  & $0.798$ {\color{gray} \scriptsize$\pm 0.009$} & $\mathbf{33.42}$ {\color{gray} \scriptsize$\pm 0.25$} & $\mathbf{0.829}$ {\color{gray} \scriptsize$\pm 0.008$} \\ \midrule
\textsc{BrainWeb} & \textsc{AAPM} & 
$32.24$ {\color{gray} \scriptsize $\pm 0.07$}  &
$0.850$ {\color{gray} \scriptsize $\pm 0.001$}  &
$29.83$ {\color{gray} \scriptsize $\pm 0.07$}  &
$0.779$ {\color{gray} \scriptsize $\pm 0.002$}  &
$\mathbf{34.58}$ {\color{gray} \scriptsize $\pm 0.17$}  &
$\mathbf{0.910}$ {\color{gray} \scriptsize $\pm 0.001$} \\ \bottomrule
\end{tabular}
\end{table*}

\subsection{Sparse-view Computed Tomography}\label{sec:sparseviewct}
Diffusion models are trained on both \textsc{AAPM} and the synthetic \textsc{Ellipses} dataset with an identical setup as in~\cite{chung2024decomposed,dhariwal2021diffusion}. These models are additionally evaluated on the \textsc{LoDoPab-CT} dataset to measure the OOD performance. For all datasets, we simulate measurements via a parallel-beam geometry using $60$ equidistant angles and $1\%$ additive relative Gaussian noise. The forward operator is implemented in ODL \cite{jonas_adler_2017_556409}. We compare SCD against RED-diff \cite{mardani2024a} and DDS~\cite{chung2024decomposed}. PSNR/SSIM values are reported in \Cref{tab:resultsSparseViewCT}, with SCD consistently outperforming RED-diff and DDS across all CT tasks, leading to a $2-4$dB higher PSNR. In \Cref{fig:sparseViewCT} we show examples of \textsc{Ellipses} to \textsc{AAPM} and \textsc{AAPM} to \textsc{Ellipses}. Additional images are shown in \Cref{fig:ct_addresults} in the appendix, including residual images. These results shows that in addition to improved PSNR/SSIM values, qualitatively visual improvements can also be observed. The \textsc{Ellipses} to \textsc{AAPM} setting is the same, which we have explored in the example in \Cref{fig:ood_sampling}. \New{Further, we evaluated a model, trained on the synthetic BrainWeb dataset, on the AAPM dataset. The BrainWeb dataset contains simulations of human heads and mimics characteristics of PET images. For this setting, we still observe significant improvement of SCD over RED-diff ($34.58$dB vs. $32.24$dB).}

\begin{figure}[]
    \centering
    \includegraphics[trim={0 0.25cm 0 0.3cm},clip,width=0.48\textwidth]{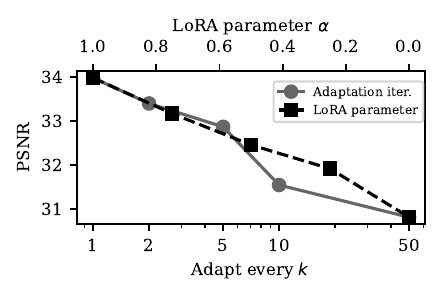}
    \caption{Ablation study on the trade-off between compute time and performance of SCD, and the effect of LoRA parameters for \textsc{Ellipses} to \textsc{AAPM} with $50$ sampling steps.} 
    \label{fig:ablation}
    \vspace{-0.1cm}
\end{figure}

\begin{figure*}[ht!]
    \centering
     \includegraphics[width=0.94\textwidth]{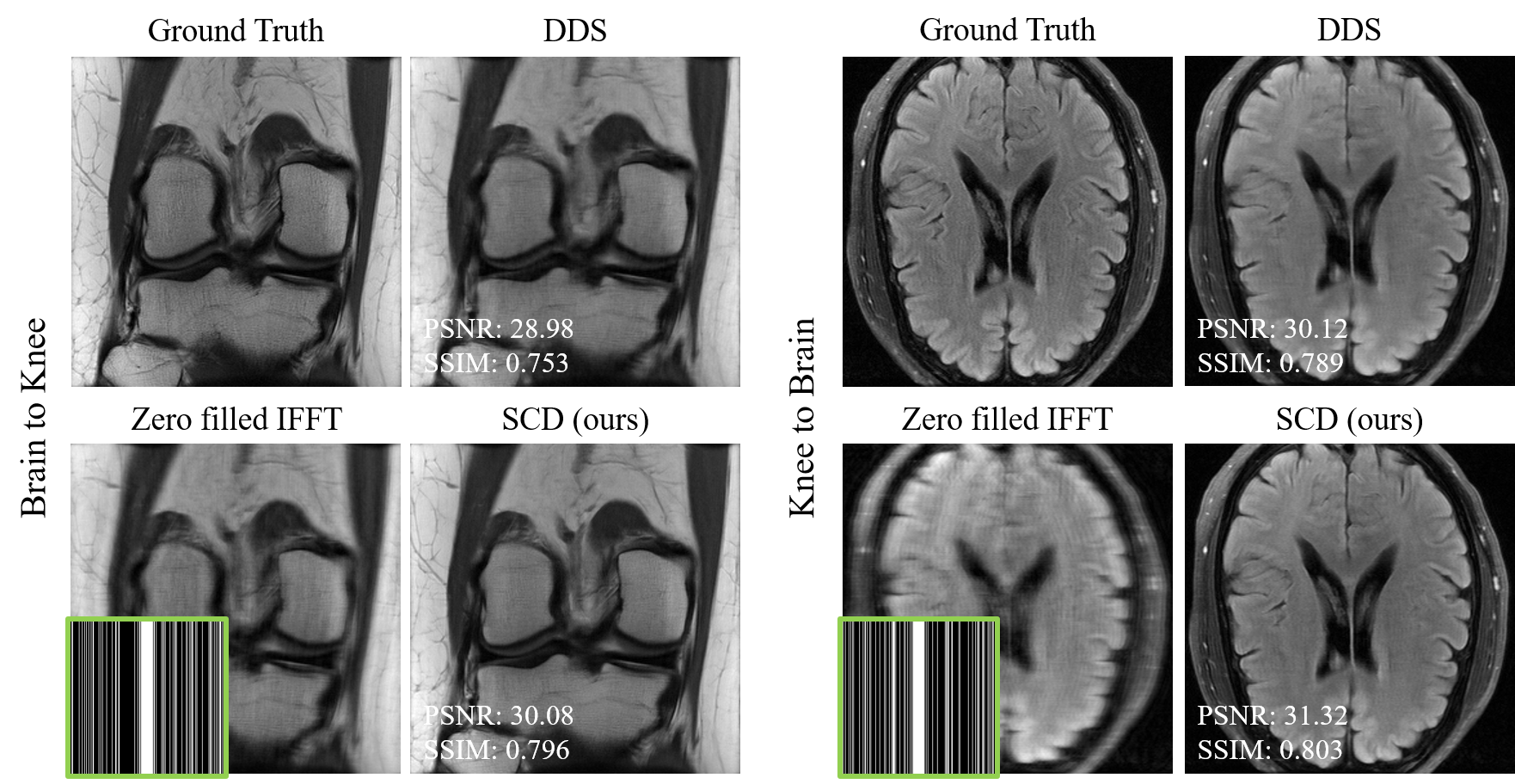}
    \caption{Results for accelerated MRI. We test two settings: training on \textsc{Brain} and testing on \textsc{Knee}, and training on \textsc{Knee} and testing on \textsc{Brain}. As a classical baseline, we also show the zero filled IFFT reconstruction.}
    \label{fig:imagesMRI}
\end{figure*}

\tabcolsep=0.11cm
\begin{table*}[ht!]
\centering
\caption{PSNR and SSIM (mean $\pm$ SE) for SCD vs. baselines on accelerated MRI. \textbf{Bold}: best}\label{tab:resultsMRI}
\begin{tabular}{llcccccccc}
\toprule  \multirow{ 2}{*}{$q(\Bx_{0})$} & \multirow{ 2}{*}{$\tilde{q}(\Bx_{0})$}  & \multicolumn{2}{c}{\textsc{DDNM}} & \multicolumn{2}{c}{\textsc{RED-diff}} & \multicolumn{2}{c}{\textsc{DDS}} & \multicolumn{2}{c}{\textsc{SCD (\textbf{ours})}} \\ 
 & & PSNR  & SSIM & PSNR  & SSIM & PSNR  & SSIM & PSNR  & SSIM       \\ \midrule
 \multirow{2}{*}{\textsc{Brain} } & \textsc{Brain}  & $28.06$ {\color{gray} \scriptsize $\pm 0.13$}  & $0.748$ {\color{gray} \scriptsize $\pm 0.006$}  & $30.07$ {\color{gray} \scriptsize $\pm 0.14$}  & $0.794$ {\color{gray} \scriptsize $\pm 0.005$}     & $30.34$ {\color{gray} \scriptsize $\pm 0.15$}  & $0.810$ {\color{gray} \scriptsize $\pm 0.006$} & \textbf{$30.47$} {\color{gray} \scriptsize $\pm 0.15$} & \textbf{$0.815$} {\color{gray} \scriptsize $\pm 0.005$} \\
& \textsc{Knee} & 
$26.98$ {\color{gray} \scriptsize $\pm 0.14$}  & 
$0.711$ {\color{gray} \scriptsize $\pm 0.005$}  & 
$28.92$ {\color{gray} \scriptsize $\pm 0.16$}  & 
$0.722$ {\color{gray} \scriptsize $\pm 0.005$}  & 
$29.47$ {\color{gray} \scriptsize $\pm 0.15$}  & $0.738$ {\color{gray} \scriptsize $\pm 0.004$} & \textbf{$30.31$} {\color{gray} \scriptsize $\pm 0.16$} & \textbf{$0.757$} {\color{gray} \scriptsize $\pm 0.005$} \\ \cmidrule{3-10} %
\multirow{ 2}{*}{\textsc{Knee} } & \textsc{Knee} & 
$28.75$ {\color{gray} \scriptsize $\pm 0.16$}  &
$0.738$ {\color{gray} \scriptsize $\pm 0.004$}  &
$30.99$ {\color{gray} \scriptsize $\pm 0.15$}  &
$0.771$ {\color{gray} \scriptsize $\pm 0.005$}  &
$31.02$ {\color{gray} \scriptsize$\pm 0.17$}  & $0.777$ {\color{gray} \scriptsize$\pm 0.005$} & \textbf{$31.10$} {\color{gray} \scriptsize$\pm 0.17$} & \textbf{$0.778$} {\color{gray} \scriptsize$\pm 0.005$}\\ 
 & \textsc{Brain} & 
 $27.01$ {\color{gray} \scriptsize $\pm 0.15$}  &
 $0.709$ {\color{gray} \scriptsize $\pm 0.005$}  &
 $28.37$ {\color{gray} \scriptsize $\pm 0.14$}  &
 $0.753$ {\color{gray} \scriptsize $\pm 0.006$}  &
 $28.62$ {\color{gray} \scriptsize$\pm 0.17$} & $0.773$ {\color{gray} \scriptsize$\pm 0.005$} & \textbf{$28.85$} {\color{gray} \scriptsize$\pm 0.18$} & \textbf{$0.780$} {\color{gray} \scriptsize$\pm 0.005$} \\ \midrule
\textsc{Ellipses} & \textsc{Brats} & 
$23.95$ {\color{gray} \scriptsize $\pm 0.21$}  &
$0.538$ {\color{gray} \scriptsize $\pm 0.008$}  &
$24.09$ {\color{gray} \scriptsize $\pm 0.22$}  &
$0.555$ {\color{gray} \scriptsize $\pm 0.009$}  &
$24.59$ {\color{gray} \scriptsize $\pm 0.23$}  &
$0.542$ {\color{gray} \scriptsize $\pm 0.008$}  &
$26.01$ {\color{gray} \scriptsize $\pm 0.22$}  &
$0.651$ {\color{gray} \scriptsize $\pm 0.007$} \\ \bottomrule
\end{tabular}
\end{table*}

For RED-diff we use $1000$ iterations as proposed in \cite{mardani2024a} and tune the regularisation parameter. For DDS we use $100$ sampling steps as proposed in \cite{chung2024decomposed} and tune the regularisation parameter and the number of CG iterations. Finally, for SCD we found that already with $50$ sampling steps and $4$ optimisation steps we were able to obtain high-quality images.

The results show that SCD is able to improve OOD performance in the sparse-view CT setting. It is worth noting that even in some in-distribution settings, i.e., $q(\Bx_{0}) = \tilde{q}(\Bx_{0})$, minor improvements are still observed, since SCD introduced the LoRA injection specific for each reconstruction. 

All sampling methods were run on a single NVIDIA GeForce RTX 4090. When comparing the computational efficiency of different sampling methods, it is evident that there are trade-offs between speed and performance. For instance, using our setup, generating a single sample with DDS takes only 6 seconds. In contrast, RED-diff and SCD require approximately 42 seconds and 48 seconds per sample, respectively.

\subsubsection{Additional Regulariser}\label{sec:add_tv}
For the sparse view CT reconstruction experiments, we utilise an additional regulariser in the fine-tuning objective. In particular, we include total variation (TV) \cite{rudin1992nonlinear}, leading to a fine-tuning objective 
\begin{align}
    \label{eq:finetuningTV}
    \Delta \boldsymbol{\theta}^{*}\!\in\!\argmin_{\Delta \theta} \{ \mathcal{L}(\Delta\boldsymbol{\theta})\!:=\!\tfrac{1}{2} \| A \Tweedie'\!-\!\By \|_2^2\!+\!\alpha_\text{TV}\!\text{TV}(\Tweedie')
    \}, 
\end{align}
which is optimised at each time step. The results are presented in Table  \ref{tab:resultsSparseViewCT_withTV}. We observe a minor performance increase on all tasks. However, this comes with the cost of having an additional hyperparameter $\alpha_\text{TV}$, governing the strength of the regularizer, which has to be set.

\begin{table}[]
\centering
\caption{PSNR and SSIM (mean $\pm$ SE) for SCD with the additional TV regulariser.}\label{tab:resultsSparseViewCT_withTV}
\begin{tabular}{llcc}
\toprule  \multirow{ 2}{*}{$q(\Bx_{0})$} & \multirow{ 2}{*}{$\tilde{q}(\Bx_{0})$}  &  \multicolumn{2}{c}{\textsc{SCD}} \\ 
 & &  PSNR  & SSIM       \\ \midrule
 \multirow{ 3}{*}{\textsc{AAPM} } & \textsc{AAPM}      & $39.77$ {\color{gray} \scriptsize $\pm 0.07$}  & $0.952$ {\color{gray} \scriptsize $\pm 0.001$}  \\
 & \textsc{LoDoPab}   & $34.88$ {\color{gray} \scriptsize $\pm 0.28$}  & $0.865$ {\color{gray} \scriptsize $\pm 0.008$} \\
 & \textsc{Ellipses}  & $35.55$ {\color{gray} \scriptsize $\pm 0.10$}  & $0.958$ {\color{gray} \scriptsize $\pm 0.001$} \\ \midrule 
\multirow{ 3}{*}{\textsc{Ellipses} } & \textsc{Ellipses} & $36.32$ {\color{gray} \scriptsize $\pm 0.09$}  & $0.969$ {\color{gray} \scriptsize $\pm 0.000$} \\
& \textsc{AAPM}  & $34.21$ {\color{gray} \scriptsize $\pm 0.12$}  & $0.890$ {\color{gray} \scriptsize $\pm 0.001$} \\ 
 & \textsc{LoDoPab} & $33.85$ {\color{gray} \scriptsize $\pm 0.24$}  & $0.845$ {\color{gray} \scriptsize $\pm 0.007$} \\ \bottomrule
\end{tabular}
\end{table}

\begin{figure*}[!ht]
    \centering
     \includegraphics[width=\textwidth]{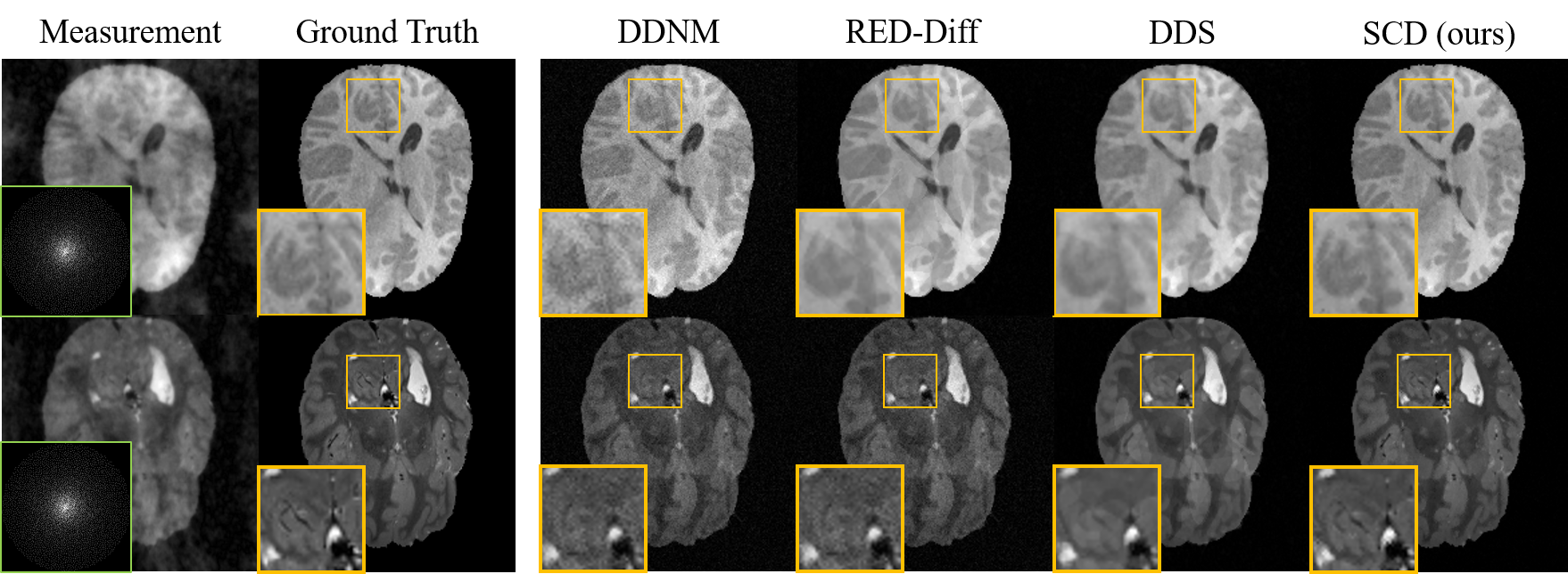}
    \caption{Results for accelerated MRI targeted for \textsc{Brats} data using the diffusion model trained only on \textsc{Ellipses} phantom.}
    \label{fig:imagesMRI_brats}
    \vspace{-0.3cm}
\end{figure*}

\subsubsection{Increase Sampling Speed}\label{sec:speedup}
\New{SCD requires adapting the injected LoRA parameters during sampling, which results in an increased sampling time. In the vanilla SCD, we adapt the diffusion model at every sampling step. Recent works reduce the sampling time by applying the data consistency update only at specific sampling steps, e.g.,~in \cite{song2024solving} the data consistency is only enforced at every $k$th sampling step. We can utilise this idea to speed-up SCD. Note that if no adaptation is performed at all, we recover the DDS algorithm. Thus, the skip $k$ can be interpreted as an interpolation between SCD and DDS. The results for \textsc{Ellipses} to \textsc{AAPM} are presented in Fig.~\ref{fig:ablation}. If we adapt at every sampling step, we recover the results reported in Table~\ref{tab:resultsSparseViewCT}. However, if we adapt every $10$th step, which increases the sampling step approximately $10$ times, we can still achieve a PSNR of $31.5$dB and beat the performance of~DDS. In addition, in Fig.~\ref{fig:ablation} we vary the parameter $\alpha$, i.e., the strength of the LoRA parametrisation. These results show that varying $\alpha$ interpolates between the performance of SCD ($\alpha=1$) and the performance of DDS ($\alpha=0$).
}
\vspace{-0.3cm}

\subsection{Accelerated MRI}
\New{We use pre-trained diffusion models on  both \textsc{KNEE} and \textsc{BRAIN} multi-coil fastMRI~\cite{zbontar2018fastmri} datasets from \cite{chung2024decomposed}.}
To simulate the measurement data with a uniform 1D sub-sampling ($\times4$ acceleration) and an $8\%$ Auto Calibrating Signal (ACS) region, we follow the original setting proposed in \cite{zbontar2018fastmri}. Additionally, we add $1\%$ relative Gaussian noise to the measured data. The coil sensitivity maps are pre-estimated using ESPiRiT~\cite{uecker2014espirit}. Following \cite{jalal2021robust, chung2024decomposed}, we use the minimum variance unbiased estimate (MVUE) images as ground truth. 

\Cref{tab:resultsMRI} shows that SCD improves DDS on OOD reconstructive settings. Adapting from \textsc{BRAIN} to \textsc{KNEE} requires capturing high-frequency image features, which are not present in \textsc{Brain}. \Cref{fig:imagesMRI} shows example reconstructions. Although the performance gains in terms of PSNR/SSIM are not as significant as for CT, we can observe that DDS introduces hallucinations in the reconstruction, which can be resolved using SCD.
To further study the effectiveness of SCD in the MRI setting, we take the diffusion model trained on the \textsc{Ellipses} and apply it to the \textsc{Brats} dataset. In Table~\ref{tab:resultsMRI}, we see the improvement achieved by SCD is even more pronounced, which can also be seen clearly in Fig.~\ref{fig:imagesMRI_brats}.

\begin{figure}[!ht]
    \centering
    \includegraphics[width=0.5\textwidth]{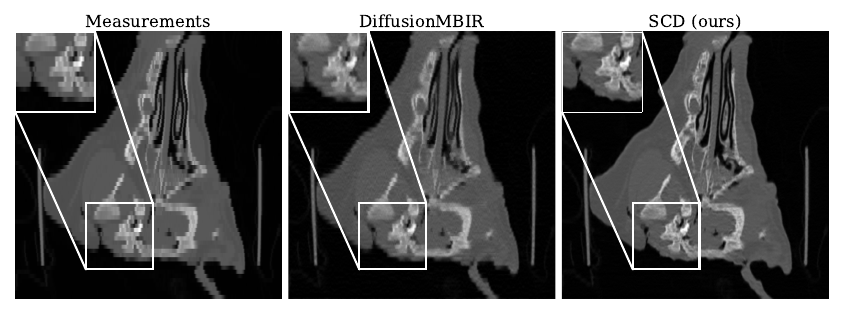}
    \caption{Coronal slice for SR. The low resolution image (left) is subsampled in $z$-direction. The DiffusionMBIR reconstruction (middle) has still visible artefacts in the $z$-direction, while SCD (right) is able to remove these artefacts.}
    \label{fig:images3DSupResSagittal}
    \vspace{-0.15cm}
\end{figure}

\begin{figure}[ht!]
    \centering
    \includegraphics[width=0.5\textwidth]{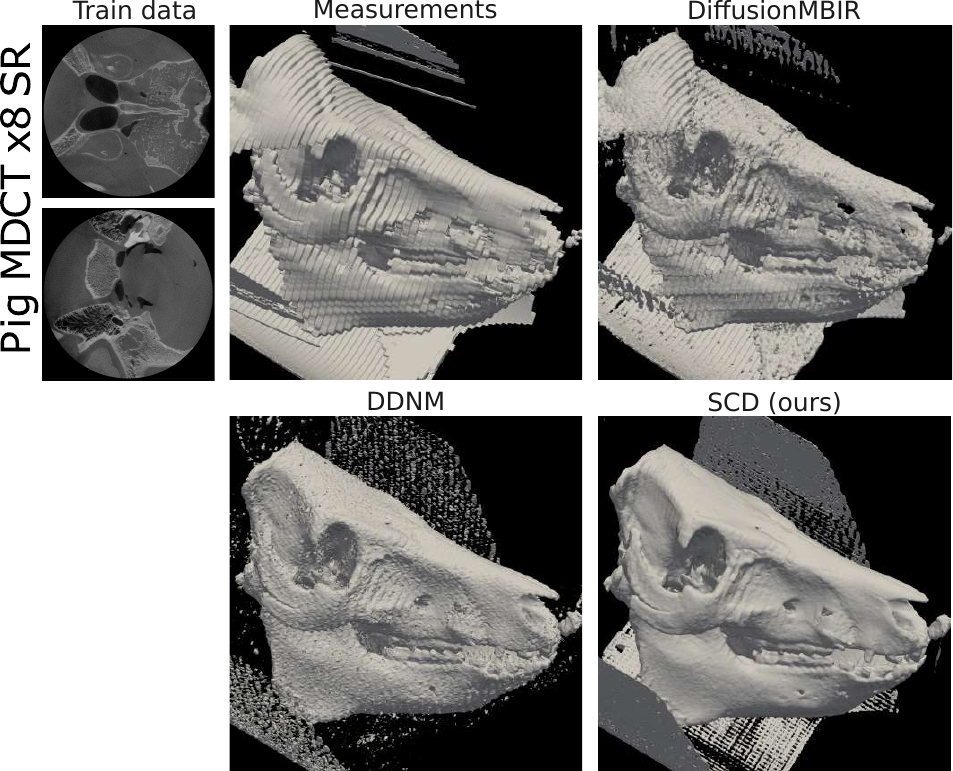}
    \caption{Left: Examples of the training data. Right: Full 3D rendering of the $\mu$CT of the head of a pig.}
    \label{fig:3Dpighead}
\end{figure}

\subsection{Super-resolution}
For this task, we employ the reconstructed $\mu$CT of the head of a pig with a reduced field-of-view (FOV), imaging the interior with a very high resolution of 0.1 mm$^3$. \Cref{fig:training_images} in the appendix shows some example images that were used to train an unconditional diffusion model. We apply SCD to enhance the vertical resolution to achieve an isotropic resolution of the 3D volume. The reconstruction task is then formulated as $\times 8$ SR on the $z-$axis. The forward operator is implemented as an average downsampling operator, i.e., taking the average of $N$ neighbouring slices. For the volumetric SR task, we benchmark against DiffusionMBIR \cite{chung2023solving}, which is a framework to apply pre-trained 2D diffusion models for 3D reconstruction and DDNM \cite{wang2022zero}. \Cref{fig:images3DSupResSagittal} shows the results on SR, where SCD greatly outperforms DiffusionMBIR. Further, in \cref{fig:3Dpighead} we provide a full 3D rendering of the upsampled pig head. It can be observed that SCD is able to provide a smoother reconstruction, which is more consistent accros the coronal axis. Here, we only provide a qualitative analysis as there is no available ground truth image to compare against.

As our pre-trained diffusion model was trained on 2D slices, it cannot be directly used for 3D reconstruction. Instead, for SCD we take we reconstruct each slice individually. Instead, one could use similar ideas as DiffusionMBIR or use a joint adaptation $\boldsymbol{\Delta \theta}$ for all slices to reduce the computational time. However, still, the slice-by-slice reconstruction is able to recover a lot of structure as is visible in the 3D visualisation of the complete up-sampled volume \cref{fig:3Dpighead}.

\section{Discussion}
\New{One concern for the practicability is the increase in the computational cost for SCD. In \ref{sec:speedup} we have provided first experiments to speed-up SCD by adapting the parameters only every $k$th sampling step. Recent work \cite{yu2023freedom} divides the diffusion process into three stages, i.e., chaotic, semantic and refinement stage, and argues that guidance in the chaotic stage is not necessary and most features are generated in the semantic stage. Thus, we could apply SCD only during the latter two stages. Improving the sampling speed of diffusion models is one research focus and we expect that a lot of the future improvements in this direction can be applied to SCD.}

\New{SCD adapts the parameters based on the likelihood of the measured data, in particular we require access to the forward operator and its adjoint. Similar to other likelihood-based methods, e.g., DIP \cite{ulyanov2018deep}, this poses the risk of over-fitting. We mitigate this risk by 1) a parameter-efficient adaptation, 2) incorporation of an additional regulariser and 3) applying early stopping. There are performance limits for all approaches based on fitting parameters using the measured data alone. In particular, for an uninformative likelihood, adaptation may fail. This is illustrated in Fig.~\ref{fig:number_of_anlges}:  SCD's performance deteriorates with fewer CT angles, which shows the challenge for likelihood-based methods.}

\section{Conclusion}
In this work, we propose Steerable Conditional Diffusion, a method that adapts diffusion models during reverse sampling, relying solely on a single measured data. 
Our experiments across diverse imaging modalities reveal that when applied to OOD reconstruction tasks, diffusion models can generate hallucinatory features from the training dataset.
Through the proposed approach, we demonstrate that adapting diffusion models drastically mitigates these artefacts.
Furthermore, we showcase that the LoRA approach not only provides a memory-efficient fine-tuning solution but is also applicable to diffusion models in reverse sampling for solving imaging inverse problems. As future research, extending our approach to scenarios involving a more extensive collection of measured data holds promise.
\vspace{0.8cm}
\section*{Appendix}

\begin{figure}[!htb]
    \centering
    \includegraphics[width=0.56\linewidth]{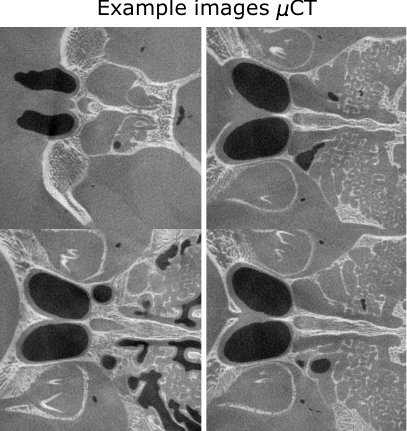}
    \caption{Example $\mu$CT images used for training the unconditional diffusion model for the super-resolution experiments.}
    \label{fig:training_images}
\end{figure}
\vspace{-0.3cm}

\begin{figure}[!htb]
    \centering
    \includegraphics[width=0.82\linewidth]{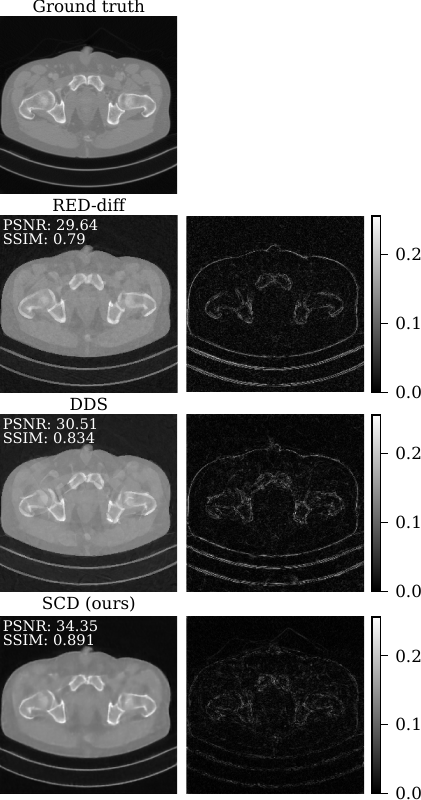}
    \caption{Additional Results for sparse-view CT for train on \textsc{Ellipses} and test on \textsc{AAPM}.}
    \label{fig:ct_addresults}
\end{figure}

\clearpage

\bibliographystyle{IEEEtran}
\bibliography{main}

\end{document}